# A Matlab Implementation of a Flat Norm Motivated Polygonal Edge Matching Method using a Decomposition of Boundary into Four 1-Dimensional Currents.

Simon Morgan, Los Alamos National Laboratory, Wotao Yin, Rice University
Kevin Vixie, Washington State University

The work of Simon Morgan and Kevin Vixie was supported by the Department of Energy through the LANL/LDRD office (Grant code: X1LJ.  LA-UR # 09-02283). The work of W. Yin was supported in part by NSF CAREER Award DMS-0748839 and ONR Grant N00014-08-1-1101

**Abstract**: We describe and provide code and examples for a polygonal edge matching method.

Our goal is to find a score to match two polygons *P1* and *P2* embedded in a rectangle *R* of the plane, of height *I* and width *J*. Using a pixel based representation of the polygon we find pixel based representations of the boundary of each polygon, with four matrices representing top, bottom left and right edges separately. We smooth with a Gaussian kernel, enabling matching of coincident edges and nearby edges. We match top edges to top edges, left edges to left edges and so on.  Not allowing cancellation between left and right edges, or between top and bottom edges, gives more sensitivity.

Polygons which represent templates and images or occluded images where only part of the original boundary is present can be matched. Polygons which differ by small deformations can be matched. The code here can be combined with registration techniques if required. Following [2], a distance function can be defined using an edge matching score. E(*P1,P2*) for the two polygons. We can define it as d(*P1,P2*)=E(*P1,P1*)+E(*P2,P2*)-2E(*P1,P2*). The capability of partial matching of nearby boundary or curves is analogous to properties the flat norm on currents. See [3] for an introduction to currents and the flat norm. Existing implementations of the flat norm [1],[2] however allow for cancellation between top and bottom edges and cancellation between left and right edges within the same polygon. This would prevent the match shown in figure 4 of the slender regions protruding to the right of the two polygons that do not intersect.

The four types of edges can also be used to represent unions of oriented curves in the plane. This can include graphs and trees. In this general setting we can consider the four matrices as a decomposition of the curves into four currents.

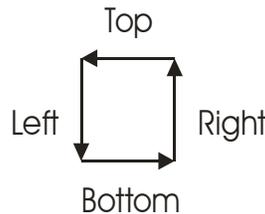

Figure 1: Correspondences between top, bottom, left and right boundary edges and oriented curves

With an appropriate sign convention as used in oriented boundary integrals with Stokes theorem, the top edges can be interpreted as horizontal components of oriented curves going to the left, bottom edges as horizontal components of oriented curve going to the right, left edges as vertical components of oriented curves pointing down, and right edges as vertical components of oriented curves pointing up. For unoriented curve matching we only make a distinction between vertical and horizontal components, requiring only two matrices; all vertical components of a curve are represented with a positive number in the vertical component matrix. This unoriented case corresponds to a decomposition into two varifolds[3], one for vertical and one for horizontal.

# The Method

**Step 1: Create pixel bitmap versions of polygons**
Convert each polygon to an *I* by *J* logical matrix (*M1* and *M2*) representing each polygon in *R*. Entry *M1(i,j)* =1 if point *(i,j)* in *R* is inside polygon *P1* and 0 if not. Similarly for *M2*.

**Step 2: Create the difference matrices of polygonal representations**
*DT1(i,j)*=1 if *M1(i,j)-M1(i-1,j)*=1, otherwise *DT(i,j)*=0. This detects upper edges of polygon pixel representations. *DT1(1,j)*=0; similarly for *DT2*.
*DB1* and *DB2* detect bottom edges.
*DL1* and *DL2* detect left edges.
*DR1* and *DR2* detect right edges.

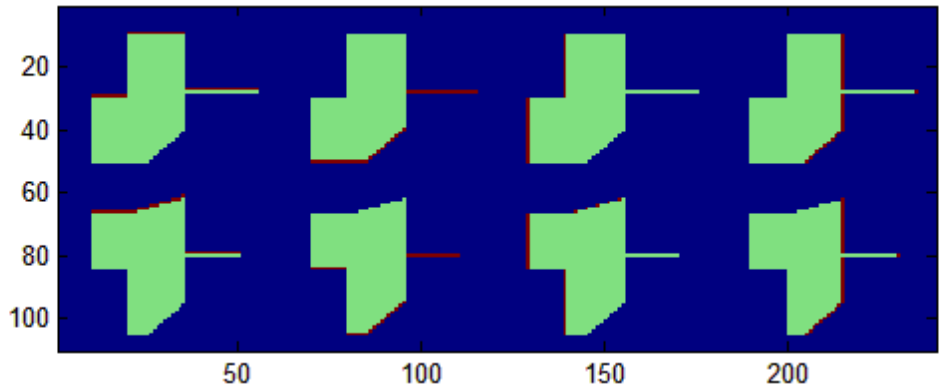

Figure 2: Four edges types for each polygon

The top row row of figure 2 shows polygon 1 and the bottom row shows polygon two. Edges types are shown in red in the order across the page; top, bottom left and right.

**Step 3: Convolve difference matrices**

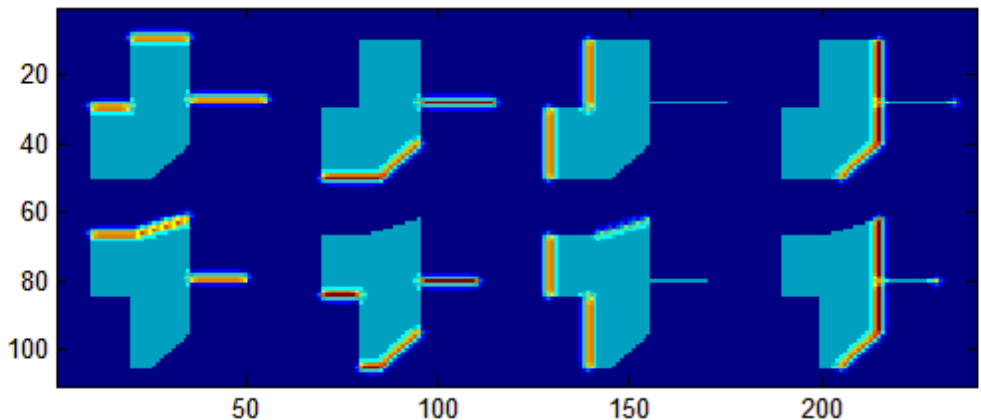

Figure 3: Four edges types convolved for each polygon

A compact Gaussian kernel is convolved with each difference matrix.
CT1=conv(DT1,Gaussian_kernel), and similarly for the others.

**Step 4: Compute Edge Matching matrices**
EMT=CT1.*CT2, represents top edge matches
EMB=CB1.*CB2, represents bottom edge matches
EML=CL1.*CL2, represents left edge matches
EMR=CR1.*CR2, represents right edge matches
These are point wise products, e.g.: EMT(i,j)=CT1(i,j)*CT(i,j).

**Step 5: Compute total score**
This uses a Euclidean metric on the net vertical and net horizontal components.
EM=((EMT+EMB)$^2$+(EML+EMR)$^2$)$^{1/2;}$
Edge_match_score=sum(sum(EM));
sum(sum(EM)) is the sum of all entries in matrix EM.

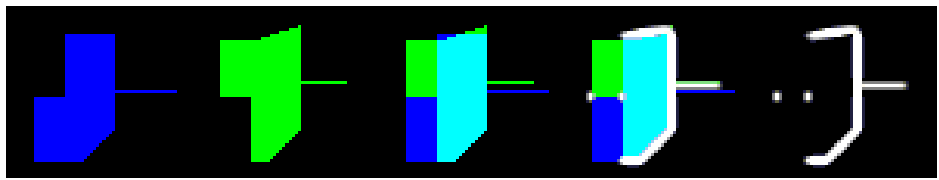

**Figure 4 (a to e): Left to right:(a) M1 (polygon 1); (b) M2 (polygon2); (c) the two polygons superimposed, (d) the two polygons superimposed with edge matching shown; and (e) EM the edge matches shown without the polygons.**

Notice that in (c) the mid left the two polygons share an edge, but for one polygon (a) this is a top edge for the other (b) it is a bottom edge. Therefore that shared edge does not match in (d) and (e). The two dots in (e) are from nearby vertical components of the polygonal edges. Notice also that the the two slender protrusions of the polygons going to the right are matched, even though they do not intersect.

## Further Reading

[1] Simon P. Morgan and Kevin R. Vixie. L$^1$TV computes the flat norm for boundaries.
arXiv:math/0612287
This paper describes the flat norm applied to boundary currents.

[2]Marc Vaillant and Joan Glaun'es. Surface matching via currents. In *Proceedings of Information Processing in Medical Imaging (IPMI 2005)*, volume 3565 of *Lecture Notes in Computer Science*. Springer, 2005.
This paper describes the use of current representations of surfaces and the use of Gaussian kernels to perform a flat norm type calculation

[3]Frank Morgan. *Geometric Measure Theory: A Beginner's Guide*. Academic Press, 3rd edition, 2000.
This gives an introduction to currents, varifolds and the flat norm.

## Appendix: Matlab code

### Sample data:

```
P1=[30 10;50 10;50 25;40 35;28.1 35;28.1 55;27.9 55;27.9 35;10 35;10 20;30 20;30 10];

P2=[12 10;29 10;29 20;50 20;50 25;40 35;25.1 35;25.1 50;24.9 50;24.9 35;7 35;12 20];
```

### M-file:

```
function Edge_match_score=BasicEdgeMatch(P1,P2)

% Note that P1 and P2 are 2 column matrices giving lists of vertex coordinates.
% All entries of P1 and P2 must be >0
% Polygons are only represented by integer lattice points and so P1 and P2 should
% be scaled accordingly. See sample data.

%%%%%%%%%%%%%%%%%%%%%%
% STEP 1
%%%%%%%%%%%%%%%%%%%%%%

x=max(max(P1(:,1)),max(P2(:,1)))+5;
y=max(max(P1(:,2)),max(P2(:,2)))+5;

X=1:x;
X=repmat(X',1,y);

Y=1:y;
Y=repmat(Y,x,1);

M1=inpolygon(X,Y,P1(:,1),P1(:,2));
M2=inpolygon(X,Y,P2(:,1),P2(:,2));

%%%%%%%%%%%%%%%%%%%%%%
% STEP 2
%%%%%%%%%%%%%%%%%%%%%%

bound1r=M1-[M1(2:end,:);M1(1,:)];
bound1c=M1-[M1(:,2:end),M1(:,1)];
bound2r=M2-[M2(2:end,:);M2(1,:)];
bound2c=M2-[M2(:,2:end),M2(:,1)];

DT1=bound1r<0;
DB1=bound1r>0;
DL1=bound1c<0;
DR1=bound1c>0;
DT2=bound2r<0;
DB2=bound2r>0;
DL2=bound2c<0;
DR2=bound2c>0;

DT1=double(DT1);
DB1=double(DB1);
DL1=double(DL1);
DR1=double(DR1);
```

```matlab
DT2=double(DT2);
DB2=double(DB2);
DL2=double(DL2);
DR2=double(DR2);

imagesc([2*DT1+M1,DB1+M1,2*DL1+M1,DR1+M1;2*DT2+M2,DB2+M2,2*DL2+M2,DR2+M2]);

%%%%%%%%%%%%%%%%%%%%%%
% STEP 3
%%%%%%%%%%%%%%%%%%%%%%

G=fspecial('gaussian',5,1)/.1621;
G=double(G);

CT1=conv2(DT1,G);
CB1=conv2(DB1,G);
CL1=conv2(DL1,G);
CR1=conv2(DR1,G);
CT2=conv2(DT2,G);
CB2=conv2(DB2,G);
CL2=conv2(DL2,G);
CR2=conv2(DR2,G);

figure
imagesc([CT1(3:end-2,3:end-2)+M1,CB1(3:end-2,3:end-2)+M1,CL1(3:end-2,3:end-2)+M1,CR1(3:end-2,3:end-2)+M1;CT2(3:end-2,3:end-2)+M2,CB2(3:end-2,3:end-2)+M2,CL2(3:end-2,3:end-2)+M2,CR2(3:end-2,3:end-2)+M2]);

%%%%%%%%%%%%%%%%%%%%%%
% STEP 4
%%%%%%%%%%%%%%%%%%%%%%

EMT=CT1.*CT2;
EMB=CB1.*CB2;
EML=CL1.*CL2;
EMR=CR1.*CR2;

figure
imagesc([EMT(3:end-2,3:end-2)+M1+M2,EMB(3:end-2,3:end-2)+M1+M2,EML(3:end-2,3:end-2)+M1+M2,EMR(3:end-2,3:end-2)+M1+M2]);

%%%%%%%%%%%%%%%%%%%%%%
% STEP 5
%%%%%%%%%%%%%%%%%%%%%%

    EM=((EMT+EMB).^2+(EML+EMR).^2).^(0.5);

    Edge_match_score=sum(sum(EM));

 figure
 imagesc([[M1,-2*M2];[(M1-2*M2),EM(3:end-2,3:end-2)]]);
 title(['Edge_match_score(P1,P2)=',num2str(Edge_match_score)])
 drawnow
 pause(0.0001)
 drawnow
 figure
 Disp1(:,:,3)=M1;
 Disp2=Disp1;
 Disp2(:,:,:)=0;
```

```
 Disp2(:,:,2)=M2;
 Dispa(:,:,3)=M1;
 Dispa(:,:,2)=M2;
 Disp(:,:,1)=EM(3:end-2,3:end-2);
 Disp(:,:,3)=Dispa(:,:,3)+ Disp(:,:,1);
 Disp(:,:,2)=Dispa(:,:,2)+ Disp(:,:,1);
Dispb(:,:,1)=EM(3:end-2,3:end-2);
Dispb(:,:,2)=EM(3:end-2,3:end-2);
Dispb(:,:,3)=EM(3:end-2,3:end-2);
 imshow([Disp1,Disp2,Dispa,Disp,Dispb])
```

## Acknowledgements

Thanks for helpful discussions to: Robert Hardt, Sarang Joshi, William Allard, Selim Esedoglu, Pete Schultz and Robert Sarracino.